\documentclass[10pt,letterpaper]{article}

\usepackage{cogsci}
\usepackage{tabularx}
\usepackage{graphicx}
\usepackage{amsmath}
\usepackage{booktabs}
\usepackage{hyperref}
\usepackage{xcolor}
\hypersetup{
    colorlinks=true,
    linkcolor=blue,
    filecolor=blue,
    urlcolor=blue,
    citecolor=blue,
}

\cogscifinalcopy

\usepackage{pslatex}
\usepackage{apacite}
\usepackage{float} 

\title{Optimal Foraging in Memory Retrieval: Evaluating Random Walks and Metropolis-Hastings Sampling in Modern Semantic Spaces}
 
\author{{\large \bf James Moore (jmoore1@mit.edu)} \\
  MIT Department of Electrical Engineering and Computer Science \\
  9.66 (Undergraduate)}

\begin{document}

\maketitle
\begin{abstract}
Human memory retrieval often resembles ecological foraging where animals search for food in a ``patchy" environment. Optimal foraging means strict adherences to the Marginal Value Thereom (MVT) in which individuals exploit a “patch” of semantically related concepts until it becomes less rewarding, then switch to a new cluster. While human behavioral data suggests foraging-like patterns in semantic fluency tasks, it is still unknown whether modern high-dimensional embedding spaces provide a sufficient representation for algorithms to closely match observed human behavior. By leveraging state-of-the-art embeddings and prior clustering and human semantic fluency data I find that random walks on these semantic embedding spaces produces results consistent with optimal foraging and the MVT. Surprisingly, introducing Metropolis-Hastings, an adaptive algorithm expected to model strategic acceptance and rejection of new clusters, does not produce results consistent with observed human behavior. These findings challenge the assumption that sophisticated sampling mechanisms inherently provide better cognitive models of memory retrieval. Instead, they highlight that appropriately structured semantic embeddings, even with minimalist sampling approaches, can produce near-optimal foraging dynamics. In doing so, my results support the perspective of Hills (2012) rather than Abbott (2015), demonstrating that modern embeddings can approximate human memory foraging without relying on complex acceptance criteria. 

\end{abstract}

\section{Introduction}
Semantic memory, the system underlying knowledge of concepts and their interrelations, has long been studied as a crucial component of human cognition. The way individuals navigate this rich conceptual space—identifying relevant items and transitioning between related ideas—can be likened to how animals forage for food in their environment. Recent work has drawn parallels between memory retrieval and optimal foraging theory (OFT), suggesting that humans move through “semantic patches” of related concepts, switching locations when the rate of meaningful retrieval begins to decrease within a patch. This analogy aligns well with findings in psychology and cognitive science that emphasize the adaptive nature of memory search and the important relationship between how semantic memory is represented and the algorithms used to retrieve information.

While the proposition that people forage in semantic space as optimally as animals forage in physical space is intriguing, a key question remains under-explored: To what extent can modern semantic embeddings and retrieval models—particularly those employing high-dimensional vector spaces and advanced generative techniques—capture these human-like foraging patterns? Current work has shown that both network-based and distributional approaches to semantic representation can be tuned to produce patterns that resemble human retrieval, yet it is not fully understood whether the state-of-the-art computational models naturally give rise to this foraging behavior without explicit task-driven adjustments. Bridging this gap requires evaluating how well these computational tools align with principles drawn from ecological animal search models, and whether they implicitly incorporate the heuristic strategies suggested by optimal foraging frameworks.

A central theoretical construct in foraging research is the Marginal Value Theorem (MVT), which provides a principled way to predict when an agent will depart a resource patch. In semantic foraging, MVT-inspired models have been applied to explain when people switch from one semantic category or thematic cluster to another. However, the application of MVT and related ecological theories to semantic tasks remains relatively untested in the new generation of embedding-based computational models. Understanding how these models behave under search constraints can deepen my insight into both the nature of semantic memory and the adequacy of the representational assumptions encoded in modern embeddings.

First, I examine whether simple random walks or Metropolis-Hastings (MH) sampling processes can reproduce human-like semantic foraging patterns using state-of-the-art semantic embeddings. By comparing these relatively basic retrieval strategies against human data, I assess how faithfully these embeddings support human memory search behaviors. Second, I investigate how augmenting embeddings with descriptive information about concepts affects their underlying structure and, consequently, the emergent retrieval dynamics. my results will shed light on how representational detail interacts with search mechanisms, potentially revealing which combinations of structure and process best approximate human foraging in semantic space.

The remainder of this paper is structured as follows. First, I outline the history of semantic spaces and human memory retrieval. Next, I detail the computational methods and the construction of my semantic embeddings, including the enhancements used to provide richer concept descriptions. Next, I present a series of experiments comparing random walks and MH sampling on these embeddings, evaluating their capacity to mimic patterns observed in human fluency and retrieval tasks. I then discuss the implications of my findings, relating them back to key theoretical principles such as MVT and considering how these insights might inform future work in cognitive modeling and representation learning. Finally, I conclude by outlining possible directions for subsequent research aimed at refining my understanding of semantic memory foraging and enhancing the fidelity of computational models in capturing human-like cognition.

\section{Related Work}
Research exploring semantic retrieval has often framed human memory search in terms of foraging behavior, drawing direct analogies between the cognitive process of accessing lexical items and the ecological strategies animals use to locate resources. Early work by Troyer, Moscovitch, and Winocur (1997) introduced the concepts of clustering and switching in semantic fluency tasks—paradigms in which participants list as many words from a given category as possible within a fixed time. Their findings indicated that people tend to produce semantically related words in bursts before making a strategic “switch” to another thematic cluster, mirroring how foragers exploit one resource patch before moving on to another when returns diminish. This perspective established a foundation for linking semantic retrieval patterns to ecologically inspired theories of search.

Building on these insights, Steyvers and Tenenbaum (2005) explored how semantic networks could be constructed to model conceptual representation and guide inferences about retrieval processes. By representing words as nodes in a graph and their semantic relationships as edges, these semantic networks provided a flexible and interpretable way to simulate memory search. Subsequent developments in representation learning—such as the use of distributional embeddings and visualization techniques (van der Maaten and Hinton, 2008)—have allowed researchers to move beyond static, hand-engineered networks. Modern embeddings trained on large corpora encode semantic similarity patterns at scale, while advanced visualization methods help reveal latent structure in high-dimensional spaces. Recent embedding models, including those developed by OpenAI (e.g., text-embedding-large-3), demonstrate strong performance on benchmarks like the Massive Text Embedding Benchmark (MTEB; Muennighoff et al., 2022), demonstrating their promise in capturing human-like patterns of semantic relatedness.

A prominent line of inquiry in modeling retrieval has involved simulating human fluency tasks with random walks and related stochastic processes on semantic representations. Hills, Jones, and Todd (2012) first applied optimal foraging theory to semantic retrieval, showing that a two-stage model combining local word–word similarity cues and global frequency-based exploration replicated human clustering and switching behavior. However, Abbott, Austerweil, and Griffiths (2015) later demonstrated that a simple random walk on a semantic network derived from human free association norms could similarly reproduce the observed foraging patterns. Their findings raised questions about the role of representations themselves: did these human-generated data points implicitly encode the structural constraints necessary for such patterns to emerge, thereby reducing the complexity required of the search algorithm?

Subsequent research has further explored how different representational choices affect retrieval dynamics. Zemla and Austerweil (2018), for example, showed that semantic networks constructed directly from human fluency data closely mirrored participants’ retrieval patterns, suggesting that network topology influenced how easily a random walk could yield human-like clustering and switching. Nematzadeh, Steyvers, and Griffiths (2016) found that networks with explicit links and small-world connectivity properties were particularly conducive to producing realistic retrieval patterns with simple random walks. Such findings align closely with Abbott et al. (2015), reinforcing the notion that representational structure can play a pivotal role in shaping emergent search behavior.

Although findings and development in representation structure led to better alignment with human behavior observations, in a response to Abbott (2015), Hills et al. (2015) argued that much of the apparent simplicity of random walk models might stem from the use of free-association networks, which embed task-specific retrieval processes directly into the representation. As a result of this response more recent work has shifted toward learned embeddings and alternative search algorithms. Rather than relying on human association norms, researchers have begun to explore how embedding models and custom retrieval policies might foster foraging-like patterns. Morales (2024), for instance, proposed an agent navigating a two-dimensional embedding space, where clusters emerge organically rather than being hand-crafted or directly derived from human norms. This agent selects items locally based on both similarity and frequency—a mechanism inspired by Hills et al. (2012)—while global switches rely only on frequency, allowing the agent to “jump” to new conceptual regions. Similarly, Zhang (2022) introduced a “semantic scent” model, which guides the agent through embeddings using probabilistic transitions that mimic human transitions between clusters.

These lines of research highlight the relationship between representation space and retrieval algorithms. Where early studies established the fundamental connection between clustering, switching, and semantic foraging, newer efforts leverage complex embedding models and adaptive search strategies to reveal how both learned and human-derived structures can facilitate or inhibit retrieval patterns.

\section{Methods}
A central aim of this study is to examine whether modern high-quality embedding models can better capture human-like semantic foraging patterns compared to earlier representations. To this end, I used a state-of-the-art embedding model provided by OpenAI, text-embedding-large-3. This model was trained on a vast, diverse corpus of text and has demonstrated improved semantic fidelity on benchmark evaluations, including the Massive Text Embedding Benchmark (MTEB; Muennighoff et al., 2022). For example, performance improvements have been noted when comparing previous generations of OpenAI embeddings (e.g., text-embedding-ada-002) to the newer text-embedding-large-3 model. On English tasks in MTEB, text-embedding-large-3 model improves from 61.0\% to 62.3\%. These gains highlight how modern embeddings offer richer and more nuanced semantic spaces than legacy models like BEAGLE (Hills et al., 2012), which relied on word–word co-occurrence statistics from a much smaller corpus.

While semantic networks (Steyvers \& Tenenbaum, 2005) have been instrumental in modeling conceptual knowledge, they carry potential downsides. Specifically, human association-based networks may embed task-specific retrieval structures directly into their topology, effectively “baking in” the retrieval process and making it easier for simple random walks to mimic human-like foraging (Hills, 2015). In contrast, high-dimensional embedding models, trained on broad and unstructured corpora, provide a more neutral semantic space that does not rely on explicit associative links. This shift in representation allows us to test whether foraging-like behaviors can emerge from simpler retrieval algorithms without presupposing any human-specific search strategies in the underlying representation.

To ground my analysis in empirical data, I focus on the semantic fluency task data collected by Hills et al. (2012), which analyzed performance from 141 participants asked to list as many animals as possible within a time limit. Drawing upon the categorization scheme from Troyer et al. (1997), 22 nonexclusive semantic categories (e.g., “African animals,” “water animals,” “beasts of burden”) were applied to participant responses. Hills et al. (2012) supplemented the original Troyer et al. list with an additional 214 animal names, resulting in a final set of 550 unique animal concepts. Each animal name provides a node in my semantic space, represented as a vector in the text-embedding-large-3 model’s 1536 dimensional embedding space.

To visualize and analyze the high-dimensional semantic representation of the 550 animals, I first computed a 550×550 cosine similarity matrix, where each entry:
\[
\text{sim}(i,j) = \frac{\mathbf{v}_i \cdot \mathbf{v}_j}{\|\mathbf{v}_i\|\|\mathbf{v}_j\|}
\]
reflects the similarity between the embeddings \(\mathbf{v}_i\) and \(\mathbf{v}_j\) of animals \(i\) and \(j\).

However, initial inspections suggested that embeddings of animal names alone did not cluster as distinctly as desired as shown in Figure 1. To improve the semantic signal and produce more coherent clusters, I supplemented each animal name with a brief descriptive sentence as demonsrated in Table 1.

\begin{table}[h!]
\centering
\begin{tabularx}{0.5\textwidth}{|l|X|}
\hline
\textbf{Animal} & \textbf{Description} \\ \hline
Elephant & An elephant is a large, intelligent mammal known for its impressive size, long trunk, and social behavior, making it one of the most recognizable and beloved animals on the planet. \\ \hline
Sloth & A sloth is a slow-moving, tree-dIlling mammal known for its unique adaptations to a life in the treetops, including long limbs, a low metabolic rate, and a diet primarily consisting of leaves. \\ \hline
Horse & A horse is a large, swift-footed mammal known for its strength, agility, and long-standing relationship with humans as a working and companion animal. \\ \hline
\end{tabularx}
\caption{Descriptions of various animals generated by OpenAI's GPT-4o model}
\label{tab:animal_descriptions}
\end{table}

By embedding these descriptive sentences, I captured more contextual detail and produced a refined similarity matrix. I then employed t-distributed Stochastic Neighbor Embedding (t-SNE; van der Maaten \& Hinton, 2008) to reduce the dimensionality for visualization. The resulting 2D maps offered a clearer view of latent clusters, approximating Troyer et al.’s (1997) categories. I also generated heatmaps of both the original and enhanced similarity matrices, as well as additive matrices (summing similarity scores across subsets of categories), to confirm improved clustering which is shown in Figure 1.

\begin{figure}[h!]
  \centering
  \includegraphics[width=\linewidth]{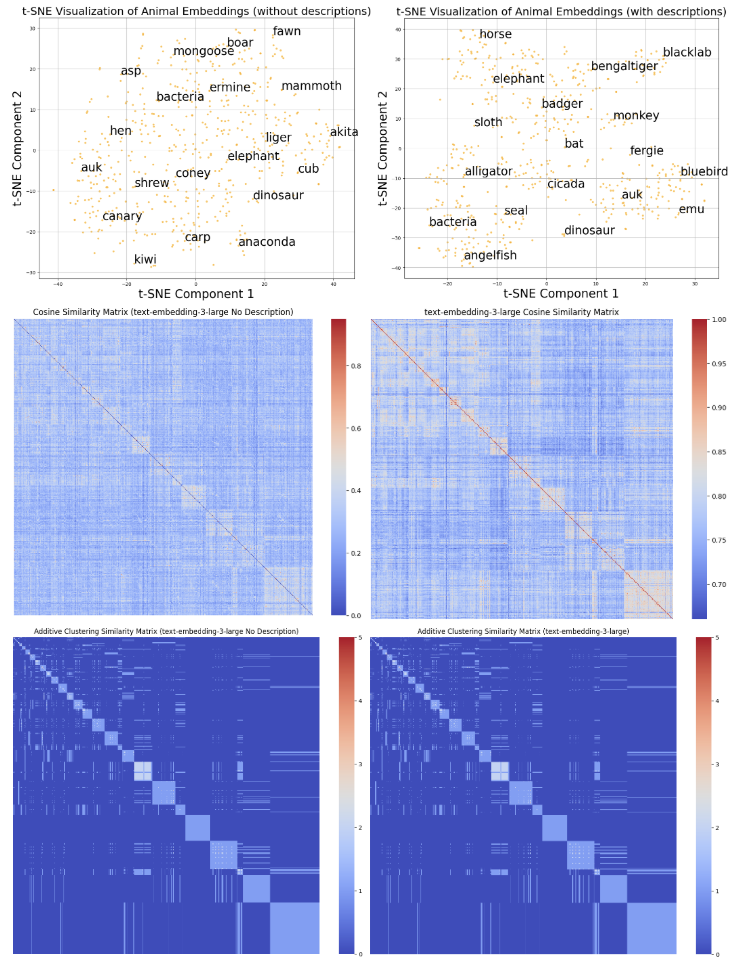}
  \caption{t-SNE, cosine similarity heatmaps and  additive clustering model visualizations. Observe that the descriptive embeddings produce better structured clusters with respect to the latent clusters.}
  \label{fig:your_label}
\end{figure}

To simulate semantic foraging, I modeled retrieval as a process unfolding over the constructed semantic space. Each animal (represented by a vector) served as a node in a fully connected graph weighted by cosine similarities. I considered two classes of retrieval models: a simple random walk and a Metropolis-Hastings (MH) sampling approach inspired by the Marginal Value Theorem (MVT).

A random walk selects the next node (animal) \(j\) from a current node \(i\) with probability proportional to their similarity:
\[
P(j | i)= \frac{\exp(\text{sim}(i,j) / T)}{\sum_{k} \exp(\text{sim}(i,k) / T)}
\]
where $T$ is a scaling parameter controlling the influence of similarity on the transition probabilities (the temperature). This model does not enforce explicit cluster-switching rules; it relies solely on the embedding-based similarities. I chose to be $T=0.027$ which produced optimal results.

To incorporate MVT-like principles, I used a Metropolis-Hastings (MH) algorithm. First, a candidate node \(j\) is sampled from a proposal distribution (e.g., uniform or similarity-based). Then, the acceptance of this candidate depends on a ratio reflecting exploitation and exploration:
\[
A(i \to j) = \min\left\{1, \frac{\pi(j)q(i|j)}{\pi(i)q(j|i)}\right\}
\]
Here, \(\pi(i)\) can be chosen to reflect semantic “profitability” (intra-cluster similarity or estimated yield of continuing to exploit a given cluster), and \(q(j|i)\) denotes the proposal probability of moving from \(i\) to \(j\). By adjusting \(\pi(i)\) to decrease as a cluster’s “value” diminishes (e.g., after retrieving several items from the same category), I simulate a departure from depleted semantic patches, aligning with the MVT’s predictions about optimal switching. The MH procedure ensures a balance between revisiting highly related items and exploring less related nodes, potentially opening pathways to new conceptual patches.

To achieve mixing and stable stationary distributions, I applied the power method, repeatedly multiplying a probability vector by the transition matrix until convergence as shown here: 
\[
\mathbf{p}_{k+1}=\mathbf{p}_k\mathbf{T}
\]
\[
\lim_{k \to \infty} \mathbf{p}_k = \mathbf{p}^* \]
where $\mathbf{p}_0$ is the uniform probability vector and $\mathbf{T}$ is the similarity matrix. This iterative approach ensures that the retrieval process reaches an equilibrium distribution reflecting both exploitation of local semantic neighborhoods and exploration of novel regions in the embedding space.

I assessed model outputs using behavioral metrics aligned with the semantic fluency literature. The inter-item response time (IRT) measures the amount of time it takes a participant to come up with two consecutive unique animals. In Hills et al. (2012), IRT is estimated to be:
   \[
   \text{IRT}(k) = \tau(k) - \tau(k-1)
   \]
Where $\tau(k)$ represents the index of a unique animal response within a list of all animal responses. For example a response list may look like: [caterpillar, salamander, salamander, mammoth, mammoth, rhino, mammoth]. In this example $IRT(3)$ would represent the response time between ``mammoth" and ``salamander" which would be $IRT(3) = 4 - 2 = 2$ since ``salamander" was mentioned twice. I measured how closely retrieved sequences adhered to Troyer et al.’s (1997) categories and how frequently the model transitioned between them. More specifically, the rate at which the model moved from one cluster to another served as an analog to foraging patch-leaving events. I expected MVT-driven models (like MH) to exhibit controlled switching patterns that mirror human behavior more closely than unconstrained random walks. One can evaluate the relevance of a random walk to the OFT by examining the last IRT in a patch to the mean global IRT. By comparing these metrics against human data, I aimed to determine whether enhanced embedding representations and theoretically grounded sampling strategies could better approximate human semantic foraging patterns.

\section{Results}

To evaluate how closely my retrieval models aligned with human semantic foraging behavior, I followed the analytical approach outlined by Abbott et al. (2015). Specifically, I compared model output to human data using two key metrics:
relative place to cluster switch IRT to mean IRT ratio (this metric captures how inter-item retrieval times vary with the position of the item relative to a cluster-switching event) and absolute deviation from last IRT in cluster to average IRT (this metric assesses how much the time to retrieve a new item deviates from the mean retrieval time within a given semantic cluster, as a function of the total number of unique animals produced). Figure 2 (adapted from Abbott, 2015) illustrates these relationships for my simulations, enabling a direct comparison between model and empirical patterns. 

\begin{figure}[h!]
  \centering
  \includegraphics[width=\linewidth]{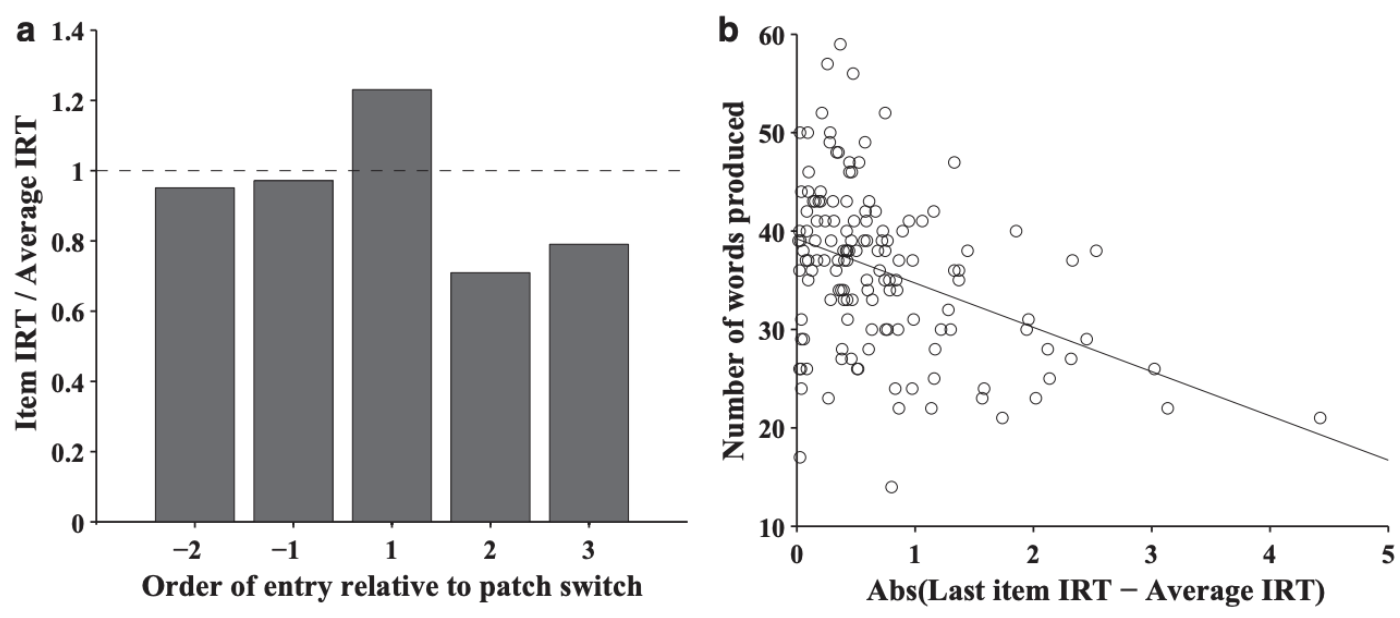}
  \caption{Empirical results from Abbott et al. (2015).}
  \label{fig:your_label}
\end{figure}

When the embeddings were derived solely from animal names, I observed limited alignment with human-like foraging patterns as show in Table 2. The random walk model, while able to retrieve items from the embedding space, showed a positive slope in the relationship between the absolute deviation from the last IRT in a cluster and the total number of unique animals produced which does not align with human behavior observations. The MH model using embeddings without descriptions did yield a significantly negative slope for the absolute deviation metric, suggesting it could mimic some aspects of human IRT patterns. However, closer inspection of the relative place to cluster switch IRT to mean IRT ratio revealed a uniform distribution across all positions, rather than the ``drop" observed in human behavior. While MH captured certain elements of the semantic foraging process, it failed to replicate the nuanced timing adjustments humans make around cluster boundaries. These results are reflected in Figure 3.

\begin{table}[h!]
\centering
\begin{tabular}{lccc}
\toprule
 & p-value & Slope & Intercept \\
\midrule
Random Walk & 1.1250e-16 & 0.3600 & 5.0383 \\
MH           & 6.8615e-04 & -21.5320 & 28.9614 \\
\bottomrule
\end{tabular}
\caption{Results for embeddings without descriptions (rounded to 4 decimal places).}
\end{table}

\begin{figure}[h!]
  \centering
  \includegraphics[width=\linewidth]{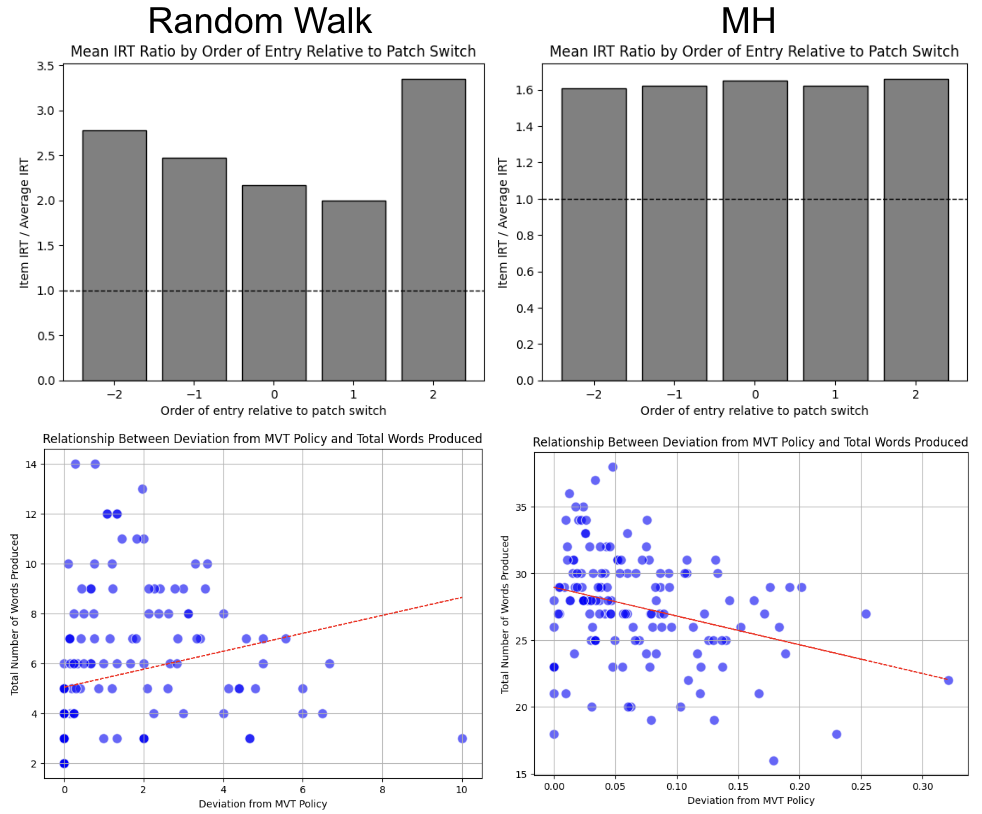}
  \caption{Results for non-descriptive embeddings.}
  \label{fig:your_label}
\end{figure}

These findings indicate that without additional semantic context, both the random walk and MH models struggled to fully approximate the complexity of human retrieval strategies.

Augmenting the animal names with descriptive sentences led to substantial improvements in model performance. Both the random walk and MH were able to achieve significant negative slopes. However, only the random walk was able to achieve qualitatively similar results within some normalization factor compared to empirical results from Abbott et al. (2015). The MH model performed similarly with descriptive and non-descriptive embeddings. These results are displayed in Table 3 and Figure 4.

\begin{table}[h!]
\centering
\begin{tabular}{lccc}
\toprule
 & p-value & Slope & Intercept \\
\midrule
Random Walk & 7.3411e-06 & -18.0037 & 39.0401 \\
MH          & 8.3803e-03 & -31.6284 & 30.3880 \\
\bottomrule
\end{tabular}
\caption{Results for embeddings with descriptions (rounded to 4 decimal places)}
\end{table}

\begin{figure}[h!]
  \centering
  \includegraphics[width=\linewidth]{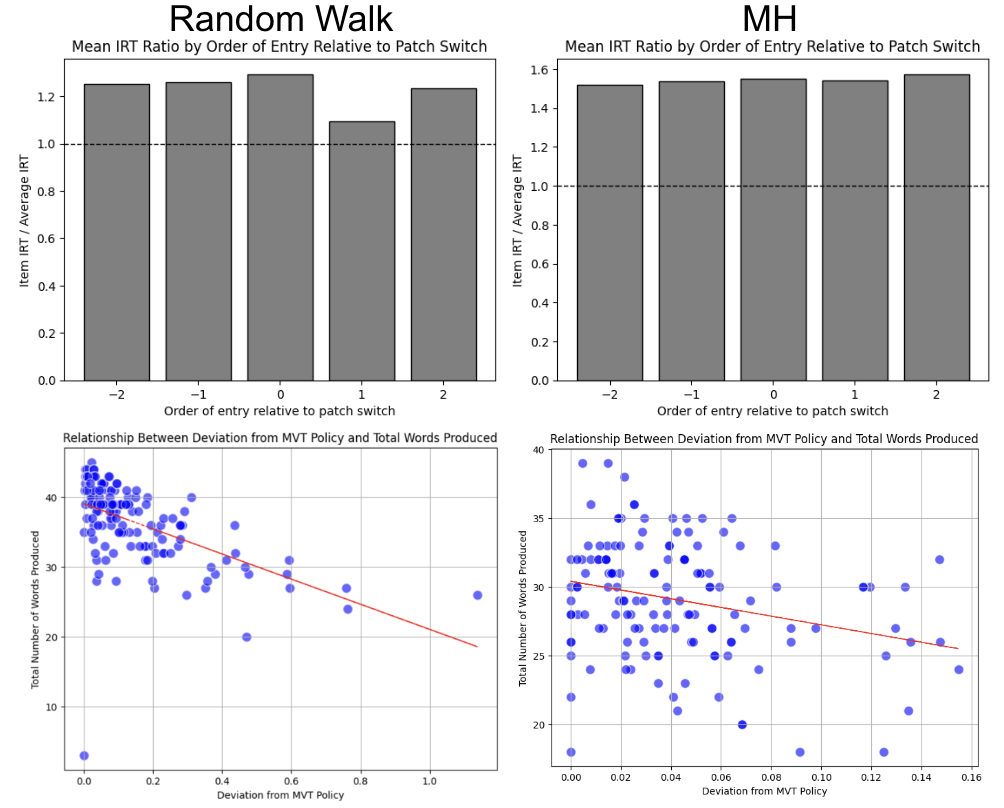}
  \caption{Results for descriptive embeddings.}
  \label{fig:your_label}
\end{figure}

\section{Discussion}

The results presented here highlight the relationship between model complexity, representational spaces, and the algorithms used for semantic retrieval. At first glance, the Metropolis-Hastings (MH) sampling approach appears to be a promising framework for simulating optimal foraging behavior in semantic memory. Its structure, inspired by the Marginal Value Theorem (MVT), offers a principled mechanism for balancing exploitation of semantically rich clusters against the exploration of new areas in conceptual space. Humans, after all, are thought to rely on accumulating experience within a given semantic “patch” and dynamically adjust their retrieval strategies based on perceived diminishing returns. This echoes MVT predictions and suggests that adaptive sampling algorithms like MH should provide a closer fit to human data than simple random walks.

Yet, my findings show a persistent gap between the theoretical appeal of MH sampling and its empirical performance. Although the MH algorithm did capture some aspects of human-like patterns—particularly when embeddings were enriched with descriptive context—it did not, upon closer inspection, replicate key nuances in the timing and patterning of cluster switches. The random walk model, by contrast, exhibited stronger alignment with human behavior, especially when operating over richly structured semantic embeddings. This raises questions about the cognitive plausibility of MH sampling’s explicit acceptance-rejection mechanics. Humans may rely less on discrete “decisions” to accept or reject a new semantic direction and more on an inherently fluid, associative retrieval process. The continuous and unfiltered nature of random walks, which simply transition to a semantically similar concept at each step, may better reflect the rapid, intuitive associations characteristic of human thought.

These observations also demonstrate the importance of the robustness of the underlying representation. Without descriptive embeddings, neither the random walk nor MH sampling aligned well with human foraging patterns; the semantic space was too sparse, and clusters too weakly defined, for either algorithm to produce the ``drop" observed in empirical data. With richer representations, the random walk emerged as a surprisingly effective proxy for human semantic search. By providing more conceptually descriptive embeddings, I effectively introduced the cluster structure necessary for associative retrieval sequences to mimic human patterns of cluster exploitation and switching. This suggests that the success of a retrieval algorithm does not hinge solely on its internal complexity or adaptive strategies, but also on the fidelity and granularity of the representational landscape it traverses.

Although my approach leverages OpenAI's state-of-the-art embedding model in text-embedding-large-3, biases inherent in the training corpus and in the embedding algorithm itself may shape the resulting semantic space in ways that differ from human conceptual organization. My experiments have also focused on a single category—animals—raising the question of how generalizable these findings are to broader domains or to richer, real-world knowledge structures. Future studies could explore larger datasets, incorporate hierarchical MH sampling procedures that allow for multi-level decisions, or introduce more cognitively plausible cues such as context or recent retrieval history. Integrating hierarchical clustering, refining acceptance functions in MH sampling, and comparing more diverse task prompts could yield insights into the conditions under which adaptive algorithms perform better than simple random walks but I leave this for a future work.

\section{Acknowledgements:}

Alvin Chen pointed me in the right direction and helped me develop this paper through its early stages. Professor Tenenbaum helped me gain an appreciation and understanding of computation cognitive science which allowed and inspired me to produce this paper.

\section{Resources:}
\url{https://github.com/jamesmoore24/optimal-foraging}

\nocite{Abbott2015}
\nocite{Hills2012}
\nocite{Troyer1997}
\nocite{Steyvers2005}
\nocite{Zemla2018}
\nocite{Morales2024}
\nocite{Nematzadeh2016}
\nocite{muennighoff2022mteb}
\nocite{Jones2015}
\nocite{vanDerMaaten2008}
\nocite{Zhang2022}

\bibliographystyle{apacite}

\setlength{\bibleftmargin}{.125in}
\setlength{\bibindent}{-\bibleftmargin}

\bibliography{CogSci_Template}

\end{document}